\documentclass{bmvc2k}

\title{SOFI: Multi-Scale Deformable Transformer for Camera Calibration with Enhanced Line Queries}

\addauthor{Sebastian Janampa }{sebasjr1966@unm.edu}{1}
\addauthor{Marios Pattichis}{pattichi@unm.edu}{1}

\addinstitution{
	Department of Electrical \& Computer Engineering\\
	The University of New Mexico\\
	New Mexico, USA
}

\usepackage{graphicx, amsmath, amssymb}
\usepackage{colortbl}
\usepackage{booktabs, dashrule, arydshln}
\usepackage{MnSymbol,utfsym}
\usepackage{dsfont}
\usepackage{multirow}

\runninghead{Sebastian Janampa and Marios Pattichis}{SOFI}

\def\etal{\emph{et al}\bmvaOneDot}

\begin{document}
	
	\maketitle
	
	\begin{abstract}		
		Camera calibration consists of estimating camera parameters such as the zenith vanishing point and horizon line. Estimating the camera parameters allows other tasks like 3D rendering, artificial reality effects, and object insertion in an image. Transformer-based models have provided promising results; however, they lack cross-scale interaction. 		
		In this work, we introduce \textit{multi-Scale defOrmable transFormer for camera calibratIon with enhanced line queries}, SOFI. 
		SOFI improves the line queries used in CTRL-C and MSCC by using both line content and line geometric features. 
		Moreover, SOFI's line queries allow transformer models to adopt the multi-scale deformable attention mechanism to promote cross-scale interaction between the feature maps produced by the backbone. 
		SOFI outperforms existing methods on the \textit {Google Street View}, \textit {Horizon Line in the Wild}, and \textit {Holicity} datasets while keeping a competitive inference speed. Code is available at: \url{https://github.com/SebastianJanampa/SOFI}
	\end{abstract}

	\section{Introduction}
	\label{sec:intro}
	Camera calibration is a fundamental task in computer vision, which enables higher-level applications such as 3D
	scene renderization, image rectification, and metrology. Estimating camera (intrinsic and extrinsic) parameters from a single-view image involves the development of methods that process the image's perspective distortions. Here, note that an example of perspective distortion is the intersection of lines on an image plane despite being parallel in the real world. 
	
	Traditional methods involve using known calibration patterns (e.g.,  chessboard images). These methods include taking multiple pictures of the 3D environment with the calibration pattern at different locations and angles. Then, corners are extracted from each image and matched using a matching technique such as  Random
	Sample Consensus (RANSAC) \cite{ransac} to compute the camera parameters. Traditional methods perform well, but require processing images of calibrated patterns and do not allow automated calibration from a single image. 
	
	Alternatively, camera calibration can be performed based on the relationship between camera parameters and geometric cues, such as vanishing points and horizon lines. We summarize these approaches based on three main steps: line segment detection, line clustering, and vanishing point estimation from each cluster. Unfortunately, such methods depend on
	man-made scenes (e.g., buildings). Their performance drops significantly on natural images or images taken in the wild. 
	
	%Finally, the last group of techniques relies on self-calibration. They take multiple pictures of the 3D environment and use multi-view geometric techniques to estimate camera parameters. Unfortunately, such methods critically depend on feature detectors that can be very sensitive to varying light conditions.
	
	Deep learning methods rely on both geometric cues and image features. In  earlier research, Deepfocal \cite{deepfocal} proposed a deep learning model to estimate the focal length of a camera based on any image taken "in the wild".  Workman \etal \cite{hlw} proposed a convolutional neural network (CNN) to predict the horizon line. Deepcalib \cite{deepcalib} proposed a CCN model to predict the intrinsic (horizon line, focal length, and distortion) camera parameters. Hold-Geoffroy \etal proposed a model that estimates both intrinsic and extrinsic at the same time. Authors in \cite{vps, parser} use geometric priors (lines) to improve estimation.
	Overall, deep learning methods based on convolutional layers can
	only capture dependencies over the support of the convolution kernels.
	Here, we note that camera calibration requires extracting information from different regions of the image that are not necessarily near each other. 
	
	More recently, Camera Calibration Transformer
	with Line-Classification (CTRL-C) \cite{ctrlc} and Multi-Scale Camera Calibration (MSCC) \cite{mscc} introduced transformer-based \cite{attention, detr} models that use the attention mechanism to extract long-term dependencies over the input images. MSCC used a coarse-to-fine technique that extended CTRL-C output queries for coarse estimation. Then, it used a second transformer model fed by a larger feature map with twice the spatial resolution of the first map.
	However, the use of a larger feature map increases inference time. At the same time, in MSCC, line geometric information is not seen by higher layers of the decoder transformer. 
	
	We propose a new neural network approach called \textit{multi-Scale defOrmable transFormer for camera calibratIon with enhanced line queries} (SOFI). Our model uses deformable attention to extract information from higher-resolution feature maps. Plus, SOFI uses line segment geometric information as input for each encoder layer. 
	
	Our contributions are summarized as follows:
	\begin{itemize}
		\item We propose a new initialization for lines' queries which  not only improves transformer-based models for camera calibration, but it also allows us to use the deformable attention mechanism.
		
		\item We reformulate the line classification functions. Additionally, we update the loss functions' coefficients, giving more importance to the camera parameters losses than to the line classification losses.
		
		\item We propose a novel framework that achieves new state-of-the-art results in in- and out- of-distribution datasets.
	\end{itemize}
	
	\section{Background}
	\label{sec: background}
	\subsection{Camera Calibration}
	Camera calibration methods are used for estimating camera parameters. Earlier methods for camera calibration are based on the use of vanishing points (VPs) and the horizon line. Schafalitzky and Zimmerman \cite{schaffalitzky}  proposed an automatic VP detection model by grouping geometric features. Tretyak \cite{tretyak} presented a parsing framework to estimate VPS by geometrically analyzing low-level geometric features to estimate the zenith VP and the horizon line. However, to be effective, these approaches require images of human-made structures (e.g., buildings), rich in geometric content that satisfy the Manhattan or Atlanta environment assumptions.
	
	On the other hand, early deep learning models directly estimated the camera parameters by extracting geometric cues using convolutional layers \cite{deepfocal,hlw}. To improve results, deep learning models used line segments detected by LSD \cite{lsd} algorithm to better understand the scene. However, these approaches require post-processing steps and have complicated architectures for integrating the line information into the deep learning model.
	
	Lee \etal \cite{ctrlc} introduced Camera calibration TRansformer with Line-Classification (CTRL-C), the first end-to-end transformer model for camera calibration. The network used a CNN backbone to extract image features and feed them into an encoder-decoder transformer model. It also utilized line information to achieve better results. Nevertheless, in CTRL-C, the input was restricted to the lowest-resolution feature map, while higher resolution feature maps were ignored. 
	Song \etal \cite{mscc} observed that low-resolution (high-level) feature maps contain global information about an image, while  higher-resolution (low-level) features capture image details. The authors \cite{mscc} propose MSCC, a multi-scale transformer model for camera calibration. MSCC uses higher-level feature maps to retrieve information about image structures and low-level feature maps to get finer details about vanishing points and horizon lines. Despite achieving new state-of-the-art (SOTA) results, standard transformers models are not time-efficient for processing low-level features because of high complexity associated with the encoder stage.
	
%	\subsection{Transformers in Computer Vision}
%	Computer vision methods have adopted transformers model for many vision tasks, achieving SOTA results. We provide a brief summary.
%	Dosovitskiy \etal \cite{vit} proposed Vision Transformer (ViT), an encoder transformers model for image classification. Swin Transformer \cite{swin_transformer} is a full transformer model that produced hierarchical feature maps that are similar to maps produced by CNN models. For vanishing point estimation, Tong \etal \cite{tong} used an encoder transformer model to classify lines in three groups, assuming Manhattan environments for the input images.
%	
%	Carion \etal introduced Detection Transformer (DETR), a novel end-to-end encoder-decoder transformer model for object detection.  Zhu \etal \cite{deformable_detr} proposed a new attention mechanism, deformable attention. Deformable DETR \cite{deformable_detr} allowed transformer models to use multi-scale feature maps and relate pixels through different level features, leading to better results for small object detection. Xu \etal \cite{letr} introduced Line Segment Detection Using Transformers without Edges (LETR), a multi-scale transformer model for line detection. In order for LETR to achieve SOTA results, it required resizing input images to ensure that the shortest size is at least 480 and at most 800 pixels while the longest is at most 1333, leading to slow inference times.
%	
	
	\subsection{DETR-based methods}
	Since Carion \etal introduced DEtection TRansformer (DETR), many methods have been proposed to improve the DETR (e.g., see  \cite{shehzadi_review}). 
%	We can classify modifications into three  categories:  backbones, queries, and  attention mechanisms. 
	In this section, we provide a short description of the DETR model and how it is used in  CTRL-C \cite{ctrlc} and MSCC \cite{mscc}. Then, we describe the deformable attention mechanism proposed in deformable DETR \cite{deformable_detr}. We end the section by discussing about the query formulation used in end-to-end transformer models.
	
	\subsubsection{Encoder in Detection Transformer for Camera Calibration}
	DEtection TRansformer \cite{detr} is the first end-to-end transformer model that discards post-processing steps. It consists of a backbone, an encoder and a decoder. The backbone produces a set of feature maps where only the highest-level feature map goes to the encoder for feature enhancing. This feature encoding consists on intra-scale processing using an attention mechanism \cite{attention} defined as:
	\begin{equation}
		\mathrm{Attn} = \mathrm{Softmax}\bigg( \frac{Q K^T}{\sqrt{d}} \bigg) V
 	\end{equation}
	where $Q$, $K$, and $V$ represent the queries, keys, and values, respectively. For the encoder part, we have that $Q=K=V=f \in \mathbb{R}^{(HW)\times d}$ where $f$ is the given feature map of height $H$, width $W$, and number of channels $d$. 
	
	CTRL-C \cite{ctrlc} and MSCC \cite{mscc} follows the previously described feature map enhancing method. The difference between the two models is that CTRL-C applies it to only the highest-level feature map, and MSCC to the two highest-level feature maps using a different encoder for each feature map. Then, each enhanced feature map is passed to a decoder to estimate the camera parameters. 
	
	In the encoder, CTRL-C estimates the camera parameters using a global feature map. Although the pixels of the feature map are rich in information, the spatial size is too small. On the other hand, MSCC uses CTRL-C output queries as pre-initilized queries to process them using a second decoder block with the second highest-level feature map. However, the global information is lost through each layer of the second decoder block. We solve this issue by using the deformable 
	attention mechanism to promote cross-scale processing while keep a good complexity time for the encoder block.
	
	\subsubsection{Deformable Attention Mechanism}
	The deformable attention mechanism is proposed in DeformableDETR \cite{deformable_detr} to allow DETR to use high-resolution feature maps without compromising inference time. The idea consists on compute $K$ sampling points as well as an attention matrix from the feature map for
	 each query $q\in Q$ such as $K\ll HW$, reducing the complexity time from $O(H^2W^2d)$ to $O(KHWd)$\footnote{Refer to \cite{deformable_detr} for the full derivations of the time complexities.}. For a single-head deformable attention mechanism, the equation is
	 \begin{equation}
	 	\mathrm{DeformAttn}(z_q, p_q, x) =  \sum_{k=1}^{K} A_{qk} \cdot x(p_q + \Delta p_{qk})
	 	\label{eq:deform_attn}
	 \end{equation}
	 
	 where $k$ indexes the sampling keys. The $k^{th}$ sampling key is computed as $p_q + \Delta p_{qk}$ where $\Delta p_{qk}$ is the sampling offset and $p_q$ is the 2d-reference point for query $q$. $A_{qk}$ is the $k^{th}$ row of the attention weight $A_q \in \mathbb{R}^{K\times d}$. The weights of $A_{qk}$ satisfy $\sum_{k=1}^{K}A_{qk} = 1$. 
	 
	 In addition to the time complexity reduction, $\mathrm{DeformAttn}$ has a variation $\mathrm{MSDeformAttn}$, where MS stands for Multi-Scale, that promotes cross-scale interaction. Similar to $\mathrm{DeformAttn}$,
	 given a set of  $L$ feature maps $\{x^l\}_{l=1}^L$, $\mathrm{MSDeformAttn}$ samples $K$ sampling point for each $x^l$. The mathematical representation for a single-head multi-scale deformable attention is 
	 \begin{equation}
	 	\sum_{l=1}^{L}\sum_{k=1}^{K}A_{lqk}\cdot x^l(\phi_l(\hat{p}_q) + \Delta p_{lqk})
	 \end{equation}
	 where $x^l\in \mathbb{R}^{H_l\times W_l \times d}$. $\hat{p}_q$ is the normalized 2d coordinates whose values lie in the range of $[0, 1]$, and that is re-scaled to dimensions of the feature map of the $l^{\text{th}}$ level by the function $\phi_l$. Like eq. \eqref{eq:deform_attn}, the attention weight $A_{lqk}$ satisfies $\sum_{l=1}^{L}\sum_{k=1}^{K}A_{lqk} = 1$.
	 
	 For the encoder, a query $q$ corresponds to a pixel in $x^l$ meaning that the pixel will interact with $K$ points from $x_l$ (intra-scale interaction), and $K$ points from the each of the remaining $(L-1)$ feature maps (cross-scale interaction). This allows the propagation of global information to low-level feature maps.

	\subsubsection{Queries in Decoder Transformers}
In this section, we summarize prior related research on modifying the query inputs for DETR-based models.  
DETR \cite{detr} uses learnable vectors to provide positional constraints. However, this works assuming that the learnable vectors contain coordinate information. Dynamic Anchor Boxes (DAB)-DETR \cite{dab} and Anchor-DETR \cite{anchor} proposed anchor boxes and points to extract positional information and improve decoder performance. 

In CTRL-C \cite{ctrlc} and MSCC \cite{mscc}, line content queries were initialized with the line geometric information, while the positional queries were initialized with zero vectors. The approaches share the first decoder layer as done in the vanilla DETR \cite{detr} model. Unfortunately, in CTRL-C and MSCC, line geometric information is lost through each layer because the positional query is zero. Also, their use of positional queries with zero vectors makes it impossible to use a deformable attention mechanism that requires the use of positional queries to extract reference points.

	\section{Methodology}
	\label{sec: methodology}
	\subsection{Model Overview}
\begin{figure}[t]
		\includegraphics[width=\linewidth]{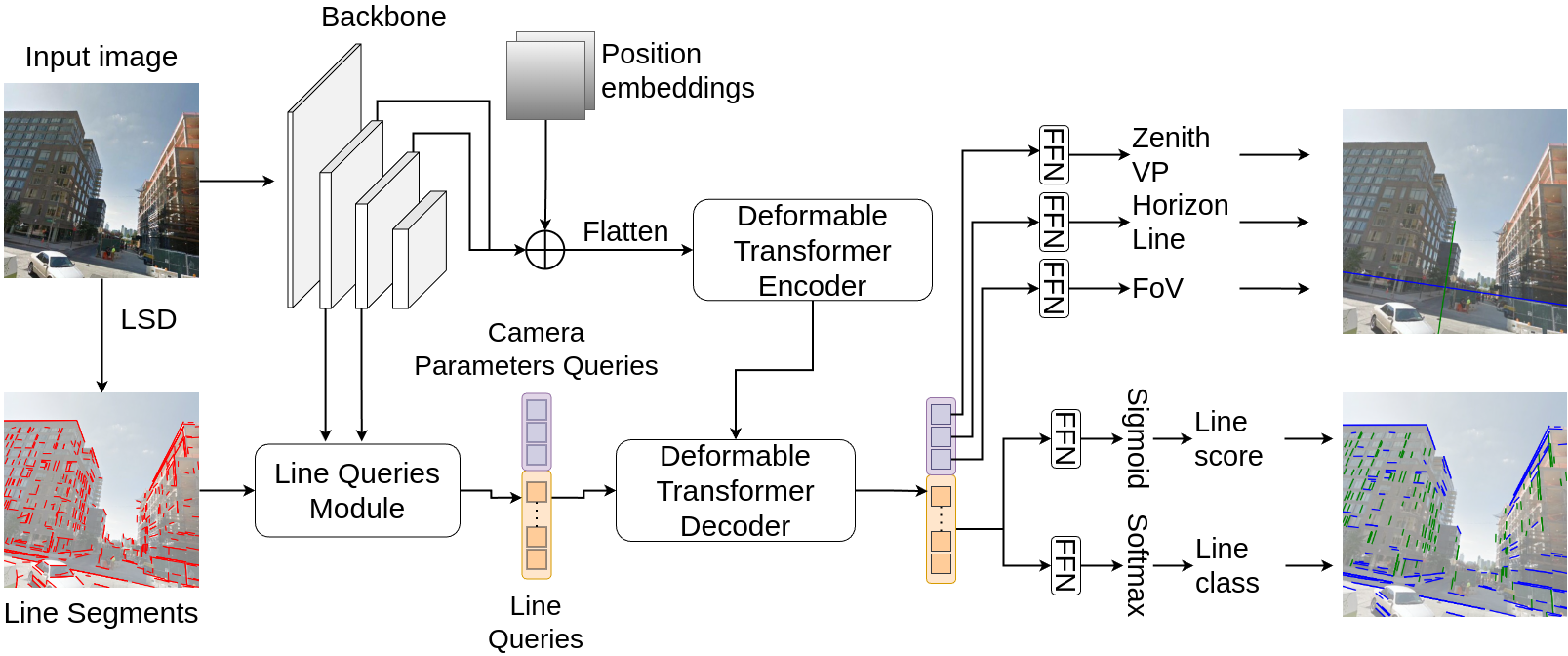}
		\caption{Model overview. Our proposed network uses ResNet50 \cite{resnet} as backbone to extract feature maps from stage 2 and 3 which are then fed to the deformable transformer encoder.}
		\label{fig: model}
	\end{figure}

We present the SOFI architecture in Fig. \ref{fig: model}. The input image is processed using ResNet50 \cite{resnet} to generate multi-scale feature maps, which are then used as encoder input and to prepare the line content queries. The encoder enhances the backbone feature maps and passes them to the decoder. The decoder outputs are processed using Feed-Forward Networks (FFNs), which produce estimates of the camera parameters (see top-right of the figure) and estimate the line segment in the image (see bottom-right of the figure).

In what follows, we provide a detailed explanation of our modified line queries and the line classification method. Our deformable encoder-decoer transformer network is adopted from Deformable DETR \cite{deformable_detr}. For our SOFI network, we use 32 sampling offsets in the deformable encoder, and 8 sampling offsets in the deformable decoder (see section 4.1 from \cite{deformable_detr} for information about the deformable attention mechanism and the sampling offsets)
	
	\subsubsection{Line Queries Module}
	Our Queries consist of content-based components $q_{con}$ and position-based components $q_{pos}$. 
	
	We follow the same procedure for position queries as in \cite{ctrlc,mscc}. Given the two endpoints of a line, we compute the homogeneous line representation $l = [a, b, c]$  where $c$ is the line offset and $a/b$ is the line slope. We also remove directional ambiguities ($l$ and $-l$ represents the same line) using  $s= [a^2, ab, b^2, bc, ac, c^2]$.
	
	Next, we define our line position-based component using: $q_{pos} = \textrm{FCL}(s)$ 
	where $\textrm{FCL}$ is a fully-connected layer that maps $s$ to a $d-$dimensionl vector, where $d$ represents the input size of the decoder.
	
	Similarly, we define our content queries using $q_{con} = \text{FCL}(\text{LS}(\mathbf{p}, \, \mathbf{F}))$
	where $\mathbf{p}$ represents the lines' endpoints, 
	$\mathbf{F}$ represents the chosen feature maps, and
	LS (for line sampling) generates $N=16$ uniformly sampled points 
	between the endpoints. 
	
	\subsubsection{Line Classification}
	\label{sec: line classification}
	We classify each detected line segment into either (i) horizontal line,
	(ii) vertical line, or (iii) other. In addition, for horizontal and vertical line classifications, we compute a confidence score that reflects
	the importance of using the detected line for camera calibration. We describe how to setup the classification loss and confidence loss functions in the section \ref{sec: loss function}. Here, we describe our line classifier and
	the regression model used for estimating the confidence level.
	
	We use a fully-connected layer followed by a sigmoid activation function for implementing line classification.
	The classifier input comes from the output of the deformable decoder.
	More precisely, our classifier is given by
	 $\mathcal{F}_{class}: \mathbb{R}^{N\times d} \rightarrow \mathbb{R}^{N\times 3}$ where $N=512$ represents the number of detected lines and $d=256$ represents the dimension of the decoder vector output for each line. 
	 
	Similarly, we use a fully-connected layer followed by sigmoid activation function for estimating the confidence score. 
	Thus, our regression network is $\mathcal{F}_{score}: \mathbb{R}^{N\times d} \rightarrow \mathbb{R}^{N\times 1}$. Here, we have the same input as for line classification. The output represents the confidence score for each one of our $N$ lines.
	
	\subsection{Loss function}
	\label{sec: loss function}
	As shown in Fig. \ref{fig: model}, we need to design loss functions
	for each output. We use: cosine distance for Zenith VP, and $L_1$ loss for
	Horizon Line and field of view (FoV) (see the Supplementary Material). We use the focal loss for
    the estimated classification probability of each class and the estimated confidence level as given by \cite{focal_loss}:
	\begin{equation}
		FL(q) = -\frac{1}{N}\sum_{i=0}^{N}(\alpha_1(1-{q}^{(i)})^\gamma\log {q}^{(i)}+\alpha_2 ({q}^{(i)})^\gamma\log (1-{q}^{(i)})),
		\label{eq: focal loss}
	\end{equation}
	where $q$ is the confidence score or the classification probability, $\alpha_1$ and $\alpha_2$ are the weighted coefficients that are determined by distances to the vanishing points as given in section 3.5 of
	\cite{ctrlc}, and $\gamma = 2$.
	
	Our overall loss $\mathcal{L}$ is defined using:
	\begin{equation}
		\mathcal{L} = 5\mathcal{L}_{zvp} + 5\mathcal{L}_{hl} + 5\mathcal{L}_{FoV} + \mathcal{L}_{score} + \mathcal{L}_{class}
	\end{equation}
	where the $\mathcal{L}_{zvp}$, $\mathcal{L}_{hl}$ and  $\mathcal{L}_{FoV}$ 
	define the loss functions of the camera parameters, while $\mathcal{L}_{class}$ and $\mathcal{L}_{score}$ are the loss functions for the confidence and the line class probability that use Eq. \eqref{eq: focal loss}. Here, we place more importance to estimating camera parameters than line classification.
	
	\section{Results}
	\label{sec: experiments}
	\subsection{Dataset}
	\textbf{Google Street View:}
	In our experiments, we train all models with 12679 images from the Google Street View dataset \cite{gsv}. The dataset consists of images of streets, buildings and landmarks that satisfy the Manhattan World assumption (three orthogonal vanishing points). Additionally, this dataset contains  535 images for validation, and 1333 for testing.\newline
	\textbf{Horizon Line in the Wild:}
	Horizon Line in the Wild (HLW) dataset \cite{hlw} only provides ground-truth for the horizon lines. We use 2019 images from this dataset to test horizon line detection in unseen data. Here, by unseen data, we refer to the fact that none of the models are trained on this dataset.\newline	
	\textbf{Holicity:}
	We test SOFI on the holicity dataset \cite{holicity} to estimate all camera parameters. The dataset consists of 2024 images. No model is trained on
	this dataset.
	
	\subsection{Implementation Details}
	\label{sec:implementation}
	We train all the models using a single Nvidia RTX A5500 GPU for 30 epochs.
For SOFI , we use a learning rate of $2\times10^{-4}$ for the first 20 epochs.
	For the last 10 epochs, we set the learning rate to $2\times10^{-5}$.
	We use AdamW for optimization with weight decay of $10^{-4}$.  
	
	We train MSCC using the recipe from LETR \cite{letr} due to the lack of training information in the original paper. First, we train the coarse stage for 30 epochs. Second, we train the coarse and fine stages for another 30 epochs. For MSCC and CTRL-C, we use a learning of $10^{-4}$, and we decrease the value to $10^{-5}$ at epoch 21.  
	
	\subsection{Comparisons}
	\begin{table}[t]\centering
		\resizebox{0.95\linewidth}{!}{%
			\begin{tabular}{lrccrccrccrcc}
				\toprule
				\multirow{2}{*}{Model} &
				\multirow{2}{*}{\phantom{a}} & \multicolumn{2}{c}{Up ($^\circ$) $\downarrow$} & \multirow{2}{*}{\phantom{a}} & \multicolumn{2}{c}{Pitch ($^\circ$) $\downarrow$} & \multirow{2}{*}{\phantom{a}} & \multicolumn{2}{c}{Roll ($^\circ$) $\downarrow$} & \multirow{2}{*}{\phantom{a}} & \multicolumn{2}{c}{FoV ($^\circ$) $\downarrow$} \\
				\cmidrule{3-4} \cmidrule{6-7} \cmidrule{9-10} \cmidrule{12-13}
				&& Mean & Med. & & Mean & Med. & & Mean & Med. & & Mean & Med. \\
				\cmidrule{3-13}
				&& \multicolumn{11}{c}{Google Street View \cite{gsv}}\\
				\cmidrule{3-13}
				Upright \cite{upright} && 3.05 & 1.92 && 2.90 & 1.80 && 6.19 & 0.43 && 9.47 & 4.42\\
				DeepHorizon \cite{hlw} && 3.58 & 3.01 && 2.76 & 2.12 && 1.78 & 1.67 && - & - \\
				Perceptual \cite{perceptual} && 2.73 & 2.13 && 2.39 & 1.78 && 0.96 & 0.66 && 4.61 & 3.89 \\
				UprightNet \cite{uprightnet} && 28.20 & 26.10 && 26.56 & 24.56 && 6.22 & 4.33 && - & - \\
				GPNet \cite{parser} && 2.12 & 1.61 && 1.92 & 1.38 && 0.75 & 0.47 && 3.59 & 2.72 \\
				CTRL-C\cite{ctrlc} && 1.71 & 1.43 && 1.52 & \textbf{1.20} && 0.57 & 0.46 && 3.38 & 2.64 \\
				MSCC \cite{mscc} && 1.75 & \textbf{1.42} && 1.56 & 1.24 && 0.58 & 0.46 && \textbf{3.04} & \textbf{2.29} \\
				SOFI (ours)&&  \textbf{1.64} & 1.44 && \textbf{1.51} & 1.28 && \textbf{0.54} & \textbf{0.43} && 3.09 & 2.79\\
				\cmidrule{3-13}
				&& \multicolumn{11}{c}{Holicity \cite{holicity}}\\
				\cmidrule{3-13}
				DeepHorizon* \cite{hlw} && 7.82 & 3.99 && 6.10 & 2.73 && 3.97 & 2.67 && -& - \\
				Perceptual* \cite{perceptual} && 7.37 & 3.29 && 6.32 & 2.86 && 3.10 & 1.82 && - & -  \\
				GPNet* \cite{parser} && 4.17 & \textbf{1.73} && \textbf{1.46} & \textbf{0.74} && 1.36 & 0.95 && - & - \\
				CTRL-C \cite{ctrlc} && 2.66 & 2.19 && 2.26 & 1.78 && 1.09 & 0.77 && 12.41 & 11.59 \\
				MSCC \cite{mscc} && 2.28 & 1.88 && 1.87 & 1.43 && \textbf{1.08} & \textbf{0.81} && 13.60 & 12.20  \\
				SOFI (ours) && \textbf{ 2.23} & 1.82 && 1.75 & 1.31 && 1.16 & 0.85 && \textbf{11.4}7 & \textbf{11.25}\\
				\bottomrule
		\end{tabular}}
		\caption{Results of camera calibration parameters on testing datasets. The $*$ is used to mark models that were trained using the SUN360 \cite{sun360} dataset.}
		\label{tab: cam params error}
	\end{table}	
	
We compare SOFI against	Upright \cite{upright}, DeepHorizon \cite{hlw}, UprightNet \cite{uprightnet}, GPNet \cite{parser}, CTRL-C\cite{ctrlc}, and MSCC \cite{mscc}. DeepHorizon, Perceptual, and GPTNet were trained on the SUN360 \cite{sun360} dataset, which is no longer available (due to license issues) for the Holicity dataset. For evaluation metrics, we use the up-vector, two angles of the rotation matrix (pitch and roll), the field of view (FoV), and the area under the curve (AUC) percentage error for the horizon line. For all the experiments, we consider a camera model with no skew (square pixels), and yaw = 0 \cite{ctrlc,mscc,hofinger}  (the third angle in the camera rotation matrix).	 For information about the estimation of the up vector, the roll, and pitch using the zenith VP and the FoV, see section 3.1 in \cite{mscc}

\begin{figure*}[t!]
		\centering
		\setlength\tabcolsep{1.5pt} % default value: 6pt
		\begin{tabular}{ccc}
		 \phantom{abc}Horizon line in the Wild & &\phantom{abc}Holicity\\
			\includegraphics[width=0.35\linewidth]{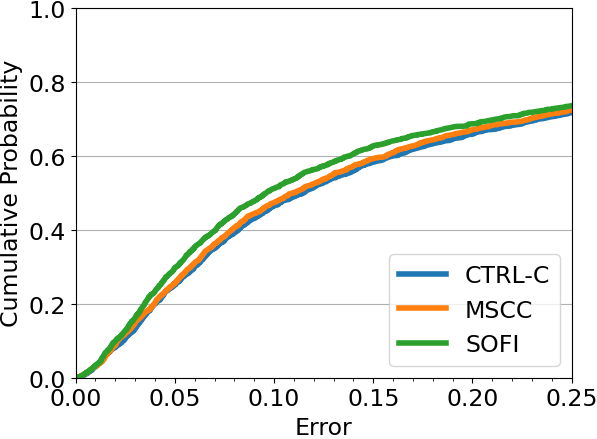} &
			\phantom{aaaaaaaaaaaa} &
			\includegraphics[width=0.35\linewidth]{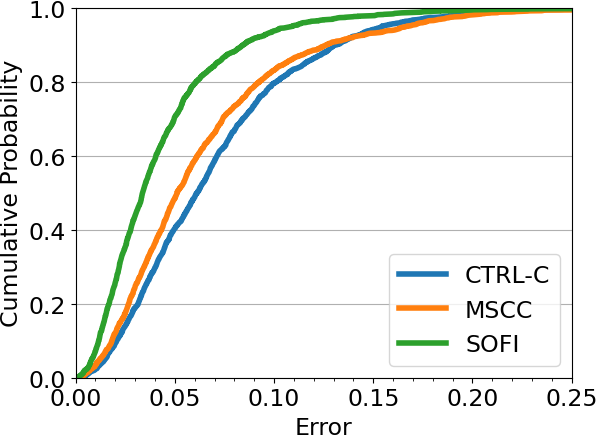} \\
		\end{tabular}
		\caption{Cumulative distribution error for the horizon line on Horizon Line in the Wild \cite{hlw} and Holicity test set \cite{holicity}.}
		\label{fig: AUC}
	\end{figure*}

We provide comparative results for the camera parameters and the horizon line in Table \ref{tab: cam params error} and Table \ref{tab: horizon line err}, respectively. SOFI provides the best results on the Google Street View dataset but for a small margin of difference. We believe this happens because the dataset is too relatively easy to learn or the training and testing sets are very similar. On the other hand, SOFI increases the margin of difference for out-of-distribution datasets (Holicity and HWL datasets. We find the most significant improvement is in the horizon line detection, where there is an improvement of 5 points with respect to MSCC, as shown in Tab. \ref{tab: horizon line err} and Fig. \ref{fig: AUC}. We want to clarify that the FoV error for the Holicity dataset is large because the sampling range of this dataset is bigger than the one in the training dataset.

	\begin{table}[t]\centering
		\resizebox{0.95\linewidth}{!}{%
			\begin{tabular}{lrcccrcccrccc}
				\toprule
				\multirow{2}{*}{Model} &
				\multirow{2}{*}{\phantom{a}} & \multicolumn{3}{c}{Google Street View \cite{gsv}} &
				\multirow{2}{*}{\phantom{a}} & \multicolumn{3}{c}{Horizon Line in the Wild \cite{hlw}} & 
				\multirow{2}{*}{\phantom{a}} & \multicolumn{3}{c}{Holicity \cite{holicity}}\\
				\cmidrule{3-5} \cmidrule{7-9} \cmidrule{11-13} && @ 0.10 & @ 0.15 & @ 0.25 && @ 0.10 & @ 0.15 & @ 0.25 && @ 0.10 & @ 0.15 & @ 0.25\\
				\cmidrule{3-5} \cmidrule{7-9} \cmidrule{11-13}
				
				DeepHorizon* \cite{hlw} && - & - & 74.25 && - & - & 45.63 && - & - & 70.13\\
				Perceptual* \cite{perceptual} && - & - & 80.40  && - & - & 38.29 && -& -&70.80\\
				GPNet* \cite{parser} && - & - & 83.12 && - & - & 48.90 && -& -& 81.72 \\
				CTRL-C && 69.49 & 78.92 & 87.16 && 24.04 & 33.56 & 46.37 && 38.84 & 55.13 & 72.31\\
				MSCC && \textbf{70.39} & 79.59 & 87.63 && 24.85 & 34.44 & 47.28 && 49.71 & 63.60 & 77.43\\
				SOFI (ours) && {70.32} & \textbf{79.84} & \textbf{87.87} && \textbf{27.93} & \textbf{37.55} & \textbf{49.69} && \textbf{59.83} & \textbf{72.05} & \textbf{82.96}\\
				\bottomrule
		\end{tabular}}
		\caption{AUC percentages (out of 100\%) for horizon line errors on testing datasets. The $*$ is used to mark models that were trained using the SUN360 \cite{sun360} dataset. @ 0.10, @ 0.20, and @ 0.25 refer to
			the area under the curve from zero to Error=0.10, 0.20, and 0.25
			as shown in Fig. \ref{fig: AUC}.}
		\label{tab: horizon line err}
	\end{table}

\begin{figure*}[t!]
		\centering
		\setlength\tabcolsep{1.5pt} % default value: 6pt
		\begin{tabular}{cccc}
			\includegraphics[width=0.22\linewidth]{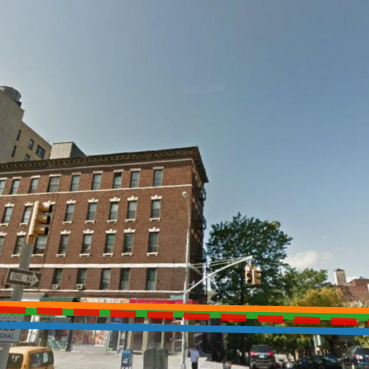} &
			\includegraphics[width=0.22\linewidth]{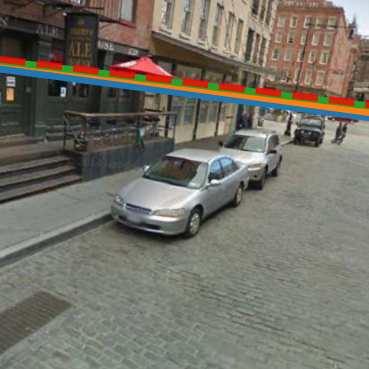} &
			\includegraphics[width=0.22\linewidth]{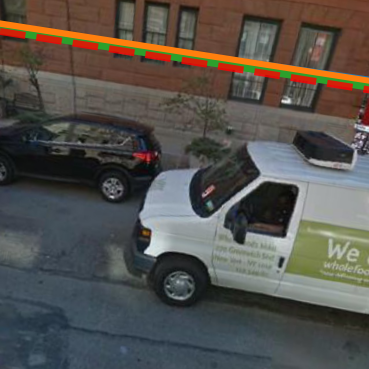} &
			\includegraphics[width=0.22\linewidth]{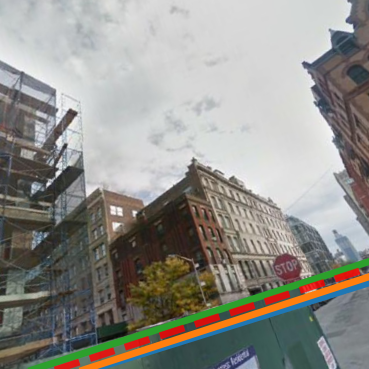} \\
			\includegraphics[width=0.22\linewidth]{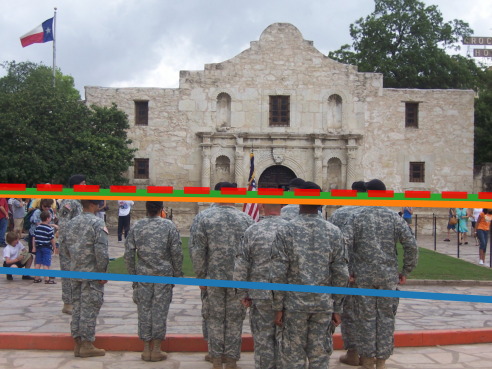} &
			\includegraphics[width=0.22\linewidth]{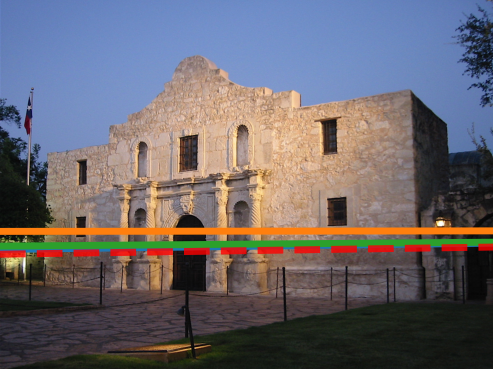} &
			\includegraphics[width=0.22\linewidth]{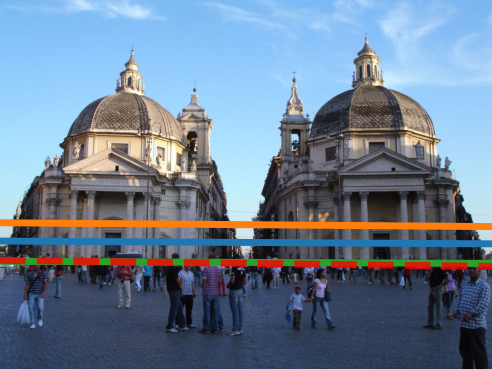} &
			\includegraphics[width=0.22\linewidth]{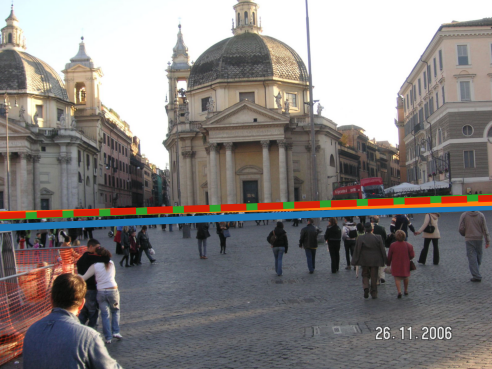} \\
			\includegraphics[width=0.22\linewidth]{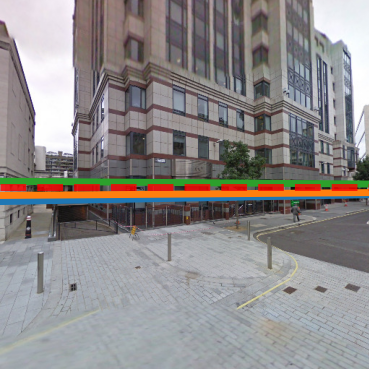} &
			\includegraphics[width=0.22\linewidth]{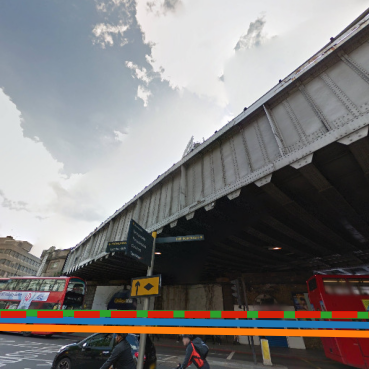} &
			\includegraphics[width=0.22\linewidth]{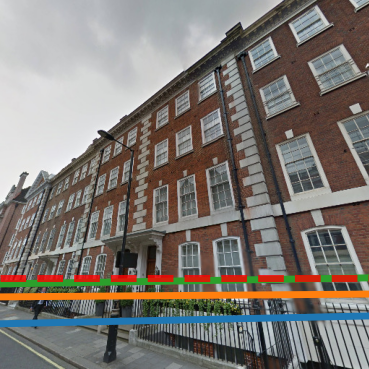} &
			\includegraphics[width=0.22\linewidth]{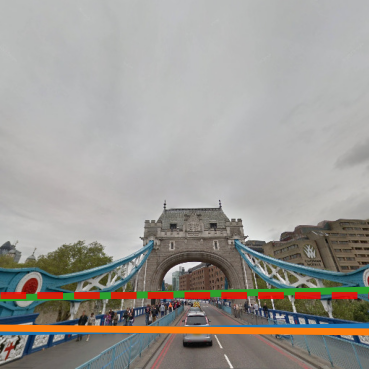} \\
		\end{tabular}
		\newcommand{\crule}[3][red]{\textcolor{#1}{\rule{#2}{#3} \rule{#2}{#3} \rule{#2}{#3} \rule{#2}{#3}}}
		{\scriptsize
			\begin{tabular}{llll}
				\crule[red]{0.01\linewidth}{0.01\linewidth} Ground Truth & 
				\crule[blue]{0.01\linewidth}{0.01\linewidth} CTRL-C \cite{ctrlc} &
				\crule[orange]{0.01\linewidth}{0.01\linewidth} MSCC \cite{mscc} &
				\crule[green]{0.01\linewidth}{0.01\linewidth} SOFI (Ours) \\
			\end{tabular}
		}
		\caption{Examples of horizon line estimation on the Google Street View \cite{gsv} test set (top row), the Horizon Line in the Wild \cite{hlw} test set (middle row), and the Holicity \cite{holicity} test set (bottom row).}
		\label{fig:horizon_line_predictions}
	\end{figure*}

    We provide horizon line estimation examples for transformer-based models in     Fig. \ref{fig:horizon_line_predictions}. We note that SOFI provides consistently accurate estimates in all of the examples. In contrast, we can see that CTRL-C and MSCC can be very inaccurate for the unseen datasets
    of the bottom two rows.
    
	We present inference speed results in table \ref{tab: speed}. 
	Despite the fact that SOFI uses multiple feature maps, at higher
	resolutions than MSCC and CTRL-C, our inference speed remains very
	competitive. SOFI runs faster than MSCC. SOFI performs nearly as fast as CTRL-C which only uses a single low-resolution feature map. 
	
\begin{table*}\centering
	\resizebox{0.95\linewidth}{!}{%
		\begin{tabular}{lrcrcrc}
			\toprule
			{Model} &
		{\phantom{a}} &{Google Street View \cite{gsv}} &
		{\phantom{a}} & {Horizon Line in the Wild \cite{hlw}} & 
		{\phantom{a}} & {Holicity \cite{holicity}}\\
			\cmidrule{3-7} 
			CTRL-C \cite{ctrlc} && \textbf{22.6}  && \textbf{19.0}  && \textbf{25.9} \\
			MSCC \cite{ctrlc} && 18.0 && 13.2 && 18.4 \\
			SOFI (ours) && 21.6 && 17.4 && 23.6\\
			\bottomrule
	\end{tabular}}
	\caption{Inference speed comparisons for transformers-based models with batch size of 1. Results are shown in terms of frames per second. Refer
	to section \ref{sec:implementation} for hardware description.}
	\label{tab: speed}
\end{table*}	

\subsection{Ablation Study}
We estimate the camera parameters using the original Deformable DETR (with four feature maps) \cite{deformable_detr} as a baseline. We modify it to use only the line geometry information to evaluate against CTRL-C \cite{ctrlc}. As shown in Tab. \ref{tab: ablation}, Deformable-DETR beats CTRL-C in the Google Street View dataset, but it does not on the Holicity dataset. This performance occurs because of the line query formulation CTRL-C has, which does not allow the estimation of proper sampling points. We refer the reader to the supplementary material for more information about this problem. There, we conducted experiments using only CTRL-C to validate the importance of choosing good coefficients for each loss and our new line queries that use geometric and content information. 

We create SOFI$^{*\dagger}$ to improve the speed and precision. SOFI$^{*\dagger}$ uses the last two feature maps and incorporates the line content information. These modifications not only improved the speed but also improved the metrics, beating CTRL-C on both datasets. Then, we replace the highest-level feature with the third highest-level one. The results are shown in SOFI$^{\dagger}$. Next, we replace the line classification proposed in \cite{ctrlc} with ours described in sec. \ref{sec: line classification}. The results in SOFI prove that our line classification module helps the model to extract richer information about line content.

We also create SOFI-Big (SOFI-B), a modified SOFI network that uses the last three feature maps. As expected, using more feature maps results in better camera parameter estimations, as reflected in the reduced errors for FoV and AUC for the horizon line. Although the significant increment in the FPS for minor accuracy improvements makes SOFI-B unsuitable for applications, it opens new borders to investigate faster cross-scale interaction methods. 

% We provide additional studies on the Supplementary Material to show the efficiency of the addition of the line content information in CTRL-C.

	\begin{table}[t]\centering
	\resizebox{0.95\linewidth}{!}{%
		\begin{tabular}{lrcrcrcrcrcrc}
			\toprule
			\multirow{2}{*}{Model} &
			\multirow{2}{*}{\phantom{a}} & {Up ($^\circ$) $\downarrow$} & \multirow{2}{*}{\phantom{a}} & {Pitch ($^\circ$) $\downarrow$} & \multirow{2}{*}{\phantom{a}} & {Roll ($^\circ$) $\downarrow$} & \multirow{2}{*}{\phantom{a}} & {FoV ($^\circ$) $\downarrow$}  & \multirow{2}{*}{\phantom{a}} & {AUC ($\%$) $\uparrow$}  & \multirow{2}{*}{\phantom{a}} & {FPS $\uparrow$} \\
			&& \multicolumn{10}{c}{Google Street View \cite{gsv}}\\
			\cmidrule{3-13}
			CTRL-C  \cite{ctrlc}&& 1.71 && 1.52 && 0.57 && 3.38 &&  87.16 && 22.6\\
			Deformable-DETR && 1.68 && 1.51 && 0.54 && 3.13 && 87.57 &&  19.6\\
			SOFI$^{*\dagger}$ && 1.69 && 1.49 && 0.61 && 3.03 && 87.48 && 22.1\\
			SOFI$^\dagger$&& 1.66 && 1.49 && 0.55 && 3.11 && 87.10 && 21.4\\
			SOFI &&  1.64  && 1.51 && 0.54 && 3.09 && 87.87 && 21.6\\ 
			SOFI-B &&  1.58  && 1.47 && 0.59 && 2.94 && 89.12 && 18.2\\ 
			\cmidrule{3-13}
			&& \multicolumn{10}{c}{Holicity \cite{holicity}}\\
			\cmidrule{3-13}
			CTRL-C \cite{ctrlc} && 2.66 && 2.26 && 1.09 && 12.41 && 72.31 && 25.9\\
			Deformable-DETR && 2.78 && 2.45 && 1.03 && 10.93 && 71.56 &&  21.0\\
			SOFI$^{*\dagger}$&& 2.52 && 2.14 && 1.07 && 12.02 && 83.18 && 25.2\\
			SOFI$^\dagger$&& 2.27 && 1.81 && 1.22 && 12.53 && 79.82 && 23.1\\
			SOFI &&  2.23  && 1.75 && 1.16 && 11.47 && 82.96 && 23.6\\
			SOFI-B &&  2.15  && 1.70 && 1.12 && 11.22 && 83.21 && 18.9\\
			\bottomrule
	\end{tabular}}
	\caption{Ablation study. $^*$ means the model uses the last two feature maps from the backbone. $^\dagger$: means the mode uses the same line classification method as in
		\cite{ctrlc}. SOFI-B uses the last three feature maps. All SOFI variations were trained with 32 and 8 sampling offsets for the encoder and decoder, respectively.}
	\label{tab: ablation}
\end{table}		

	\section{Conclusions}
\label{sec: conclusions}
We present a new model that uses deformable attention to produce new SOTA results while operating at competitive inference speed.
Our model also introduces a new line query approach that provides better camera calibration estimates than what can be achieved with DETR architectures. Our experiments document significant performance improvements in unseen datasets (Holicity and Horizon line in the Wild datasets). 

In future work, we will consider a deeper study of the encoder because of its important role in feature maps enhancing. An important observation from this work is that increasing the number of sampling points in the deformable encoder produces better feature maps. We will consider increasing the number of points without compromising speed. Another direction is to explore different cross-scale interaction mechanism that lead to an acceptable time complexity. 

\newpage

\section*{Acknowledgment}
This work was supported in part by the National Science Foundation under Grant 1949230, Grant 1842220, and Grant 1613637
	
\bibliography{paper}

\begin{thebibliography}{27}
\providecommand{\natexlab}[1]{#1}
\providecommand{\url}[1]{\texttt{#1}}
\expandafter\ifx\csname urlstyle\endcsname\relax
  \providecommand{\doi}[1]{doi: #1}\else
  \providecommand{\doi}{doi: \begingroup \urlstyle{rm}\Url}\fi

\bibitem[gsv()]{gsv}
Google street view images api.
\newblock URL \url{https://developers.google.com/maps/}.

\bibitem[Bogdan et~al.(2018)Bogdan, Eckstein, Rameau, and Bazin]{deepcalib}
Oleksandr Bogdan, Viktor Eckstein, Francois Rameau, and Jean-Charles Bazin.
\newblock Deepcalib: a deep learning approach for automatic intrinsic
  calibration of wide field-of-view cameras.
\newblock In \emph{Proceedings of the 15th ACM SIGGRAPH European Conference on
  Visual Media Production}, 2018.

\bibitem[Carion et~al.(2020)Carion, Massa, Synnaeve, Usunier, Kirillov, and
  Zagoruyko]{detr}
Nicolas Carion, Francisco Massa, Gabriel Synnaeve, Nicolas Usunier, Alexander
  Kirillov, and Sergey Zagoruyko.
\newblock End-to-end object detection with transformers.
\newblock In \emph{European conference on computer vision}, pages 213--229.
  Springer, 2020.

\bibitem[Fischler and Bolles(1981)]{ransac}
Martin~A Fischler and Robert~C Bolles.
\newblock Random sample consensus: a paradigm for model fitting with
  applications to image analysis and automated cartography.
\newblock \emph{Communications of the ACM}, 24\penalty0 (6):\penalty0 381--395,
  1981.

\bibitem[He et~al.(2016)He, Zhang, Ren, and Sun]{resnet}
Kaiming He, Xiangyu Zhang, Shaoqing Ren, and Jian Sun.
\newblock Deep residual learning for image recognition.
\newblock In \emph{Proceedings of the IEEE conference on computer vision and
  pattern recognition}, pages 770--778, 2016.

\bibitem[Hofinger et~al.(2020)Hofinger, Bul{\`o}, Porzi, Knapitsch, Pock, and
  Kontschieder]{hofinger}
Markus Hofinger, Samuel~Rota Bul{\`o}, Lorenzo Porzi, Arno Knapitsch, Thomas
  Pock, and Peter Kontschieder.
\newblock Improving optical flow on a pyramid level.
\newblock In \emph{European Conference on Computer Vision}, pages 770--786.
  Springer, 2020.

\bibitem[Hold-Geoffroy et~al.(2018)Hold-Geoffroy, Sunkavalli, Eisenmann,
  Fisher, Gambaretto, Hadap, and Lalonde]{perceptual}
Yannick Hold-Geoffroy, Kalyan Sunkavalli, Jonathan Eisenmann, Matthew Fisher,
  Emiliano Gambaretto, Sunil Hadap, and Jean-Fran{\c{c}}ois Lalonde.
\newblock A perceptual measure for deep single image camera calibration.
\newblock In \emph{Proceedings of the IEEE Conference on Computer Vision and
  Pattern Recognition}, pages 2354--2363, 2018.

\bibitem[Lee et~al.(2013)Lee, Shechtman, Wang, and Lee]{upright}
Hyunjoon Lee, Eli Shechtman, Jue Wang, and Seungyong Lee.
\newblock Automatic upright adjustment of photographs with robust camera
  calibration.
\newblock \emph{IEEE transactions on pattern analysis and machine
  intelligence}, 36\penalty0 (5):\penalty0 833--844, 2013.

\bibitem[Lee et~al.(2020)Lee, Sung, Lee, and Kim]{parser}
Jinwoo Lee, Minhyuk Sung, Hyunjoon Lee, and Junho Kim.
\newblock Neural geometric parser for single image camera calibration.
\newblock In \emph{Computer Vision--ECCV 2020: 16th European Conference,
  Glasgow, UK, August 23--28, 2020, Proceedings, Part XII 16}, pages 541--557.
  Springer, 2020.

\bibitem[Lee et~al.(2021)Lee, Go, Lee, Cho, Sung, and Kim]{ctrlc}
Jinwoo Lee, Hyunsung Go, Hyunjoon Lee, Sunghyun Cho, Minhyuk Sung, and Junho
  Kim.
\newblock Ctrl-c: Camera calibration transformer with line-classification.
\newblock In \emph{Proceedings of the IEEE/CVF International Conference on
  Computer Vision}, pages 16228--16237, 2021.

\bibitem[Lin et~al.(2017)Lin, Goyal, Girshick, He, and Doll{\'a}r]{focal_loss}
Tsung-Yi Lin, Priya Goyal, Ross Girshick, Kaiming He, and Piotr Doll{\'a}r.
\newblock Focal loss for dense object detection.
\newblock In \emph{Proceedings of the IEEE international conference on computer
  vision}, pages 2980--2988, 2017.

\bibitem[Liu et~al.(2022)Liu, Li, Zhang, Yang, Qi, Su, Zhu, and Zhang]{dab}
Shilong Liu, Feng Li, Hao Zhang, Xiao Yang, Xianbiao Qi, Hang Su, Jun Zhu, and
  Lei Zhang.
\newblock {DAB}-{DETR}: Dynamic anchor boxes are better queries for {DETR}.
\newblock In \emph{International Conference on Learning Representations}, 2022.
\newblock URL \url{https://openreview.net/forum?id=oMI9PjOb9Jl}.

\bibitem[Schaffalitzky and Zisserman(2000)]{schaffalitzky}
Frederik Schaffalitzky and Andrew Zisserman.
\newblock Planar grouping for automatic detection of vanishing lines and
  points.
\newblock \emph{Image and Vision Computing}, 18\penalty0 (9):\penalty0
  647--658, 2000.

\bibitem[Shehzadi et~al.(2023)Shehzadi, Hashmi, Stricker, and
  Afzal]{shehzadi_review}
Tahira Shehzadi, Khurram~Azeem Hashmi, Didier Stricker, and Muhammad~Zeshan
  Afzal.
\newblock Object detection with transformers: A review, 2023.

\bibitem[Song et~al.(2024)Song, Kang, Moteki, Suzuki, Kobayashi, and Tan]{mscc}
Xu~Song, Hao Kang, Atsunori Moteki, Genta Suzuki, Yoshie Kobayashi, and Zhiming
  Tan.
\newblock Mscc: Multi-scale transformers for camera calibration.
\newblock In \emph{Proceedings of the IEEE/CVF Winter Conference on
  Applications of Computer Vision (WACV)}, pages 3262--3271, January 2024.

\bibitem[Tretyak et~al.(2012)Tretyak, Barinova, Kohli, and Lempitsky]{tretyak}
Elena Tretyak, Olga Barinova, Pushmeet Kohli, and Victor Lempitsky.
\newblock Geometric image parsing in man-made environments.
\newblock \emph{International Journal of Computer Vision}, 97:\penalty0
  305--321, 2012.

\bibitem[Vaswani et~al.(2017)Vaswani, Shazeer, Parmar, Uszkoreit, Jones, Gomez,
  Kaiser, and Polosukhin]{attention}
Ashish Vaswani, Noam Shazeer, Niki Parmar, Jakob Uszkoreit, Llion Jones,
  Aidan~N Gomez, \L~ukasz Kaiser, and Illia Polosukhin.
\newblock Attention is all you need.
\newblock In I.~Guyon, U.~Von Luxburg, S.~Bengio, H.~Wallach, R.~Fergus,
  S.~Vishwanathan, and R.~Garnett, editors, \emph{Advances in Neural
  Information Processing Systems}, volume~30. Curran Associates, Inc., 2017.
\newblock URL
  \url{https://proceedings.neurips.cc/paper_files/paper/2017/file/3f5ee243547dee91fbd053c1c4a845aa-Paper.pdf}.

\bibitem[von Gioi et~al.(2010)von Gioi, Jakubowicz, Morel, and Randall]{lsd}
R.G. von Gioi, J.~Jakubowicz, J.-M. Morel, and G.~Randall.
\newblock {LSD}: A fast line segment detector with a false detection control.
\newblock \emph{{IEEE} Transactions on Pattern Analysis and Machine
  Intelligence}, 32\penalty0 (4):\penalty0 722--732, apr 2010.
\newblock \doi{10.1109/tpami.2008.300}.

\bibitem[Wang et~al.(2022)Wang, Zhang, Yang, and Sun]{anchor}
Yingming Wang, Xiangyu Zhang, Tong Yang, and Jian Sun.
\newblock Anchor detr: Query design for transformer-based detector.
\newblock In \emph{Proceedings of the AAAI Conference on Artificial
  Intelligence}, volume~36, pages 2567--2575, 2022.

\bibitem[Workman et~al.(2015)Workman, Greenwell, Zhai, Baltenberger, and
  Jacobs]{deepfocal}
Scott Workman, Connor Greenwell, Menghua Zhai, Ryan Baltenberger, and Nathan
  Jacobs.
\newblock Deepfocal: A method for direct focal length estimation.
\newblock In \emph{2015 IEEE International Conference on Image Processing
  (ICIP)}, pages 1369--1373, 2015.
\newblock \doi{10.1109/ICIP.2015.7351024}.

\bibitem[Workman et~al.(2016)Workman, Zhai, and Jacobs]{hlw}
Scott Workman, Menghua Zhai, and Nathan Jacobs.
\newblock Horizon lines in the wild.
\newblock In \emph{{British Machine Vision Conference (BMVC)}}, 2016.

\bibitem[Xian et~al.(2019)Xian, Li, Fisher, Eisenmann, Shechtman, and
  Snavely]{uprightnet}
Wenqi Xian, Zhengqi Li, Matthew Fisher, Jonathan Eisenmann, Eli Shechtman, and
  Noah Snavely.
\newblock Uprightnet: geometry-aware camera orientation estimation from single
  images.
\newblock In \emph{Proceedings of the IEEE/CVF International Conference on
  Computer Vision}, pages 9974--9983, 2019.

\bibitem[Xiao et~al.(2012)Xiao, Ehinger, Oliva, and Torralba]{sun360}
Jianxiong Xiao, Krista~A Ehinger, Aude Oliva, and Antonio Torralba.
\newblock Recognizing scene viewpoint using panoramic place representation.
\newblock In \emph{2012 IEEE Conference on Computer Vision and Pattern
  Recognition}, pages 2695--2702. IEEE, 2012.

\bibitem[Xu et~al.(2021)Xu, Xu, Cheung, and Tu]{letr}
Yifan Xu, Weijian Xu, David Cheung, and Zhuowen Tu.
\newblock Line segment detection using transformers without edges.
\newblock In \emph{Proceedings of the IEEE/CVF Conference on Computer Vision
  and Pattern Recognition}, pages 4257--4266, 2021.

\bibitem[Zhai et~al.(2016)Zhai, Workman, and Jacobs]{vps}
Menghua Zhai, Scott Workman, and Nathan Jacobs.
\newblock Detecting vanishing points using global image context in a
  non-manhattan world.
\newblock In \emph{Proceedings of the IEEE Conference on Computer Vision and
  Pattern Recognition}, pages 5657--5665, 2016.

\bibitem[Zhou et~al.(2020)Zhou, Huang, Dai, Luo, Chen, and Ma]{holicity}
Yichao Zhou, Jingwei Huang, Xili Dai, Linjie Luo, Zhili Chen, and Yi~Ma.
\newblock {HoliCity}: A city-scale data platform for learning holistic {3D}
  structures.
\newblock 2020.
\newblock arXiv:2008.03286 [cs.CV].

\bibitem[Zhu et~al.(2020)Zhu, Su, Lu, Li, Wang, and Dai]{deformable_detr}
Xizhou Zhu, Weijie Su, Lewei Lu, Bin Li, Xiaogang Wang, and Jifeng Dai.
\newblock Deformable detr: Deformable transformers for end-to-end object
  detection.
\newblock \emph{arXiv preprint arXiv:2010.04159}, 2020.

\end{thebibliography}
\newpage

\appendix

\section{Line Queries: Content and Geometric Information}

Queries are crucial in the transformer decoder module because they interact with the enhanced feature maps to produce a task-specific output. For camera parameters, transformer-based models have two types of queries: camera parameter queries $\mathbf{q}_{\text{camera}}$ and line queries $\mathbf{q}_{\text{line}}$. The camera parameters queries produce three outputs: the zenith vanishing point, the field of view, and the horizon line. The line queries classify if a line passes through the zenith vanishing point or the horizon line. 

\begin{figure}[!h]\centering
	\resizebox{\linewidth}{!}{
		\begin{tabular}{cccc}
			\includegraphics[width=0.5\textheight]{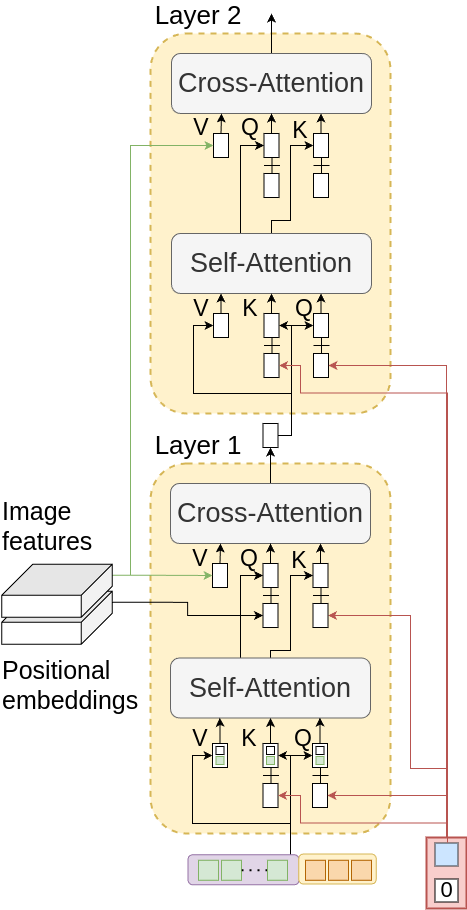} &
			\includegraphics[width=0.5\textheight]{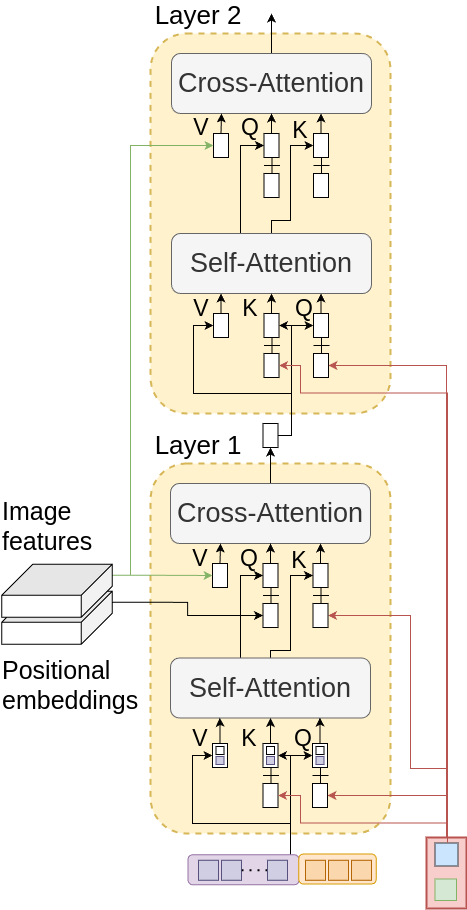} &
			\includegraphics[width=0.5\textheight]{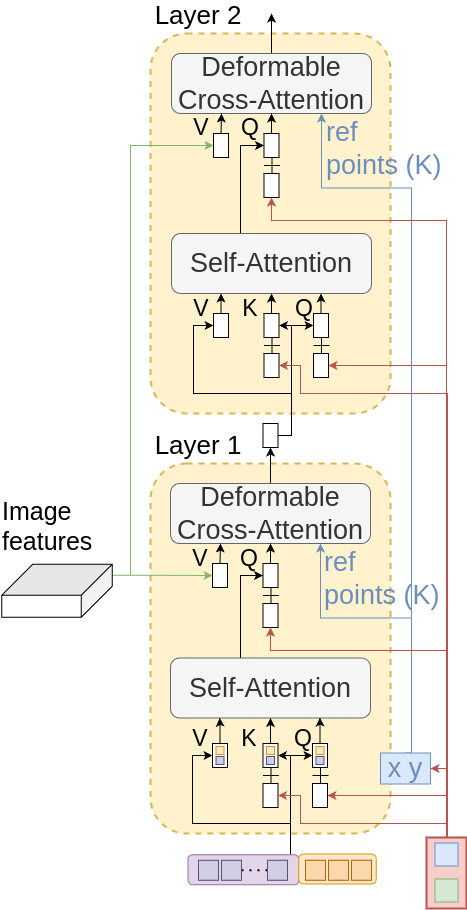} & \phantom{aaa}
			\includegraphics[width=0.25\textheight]{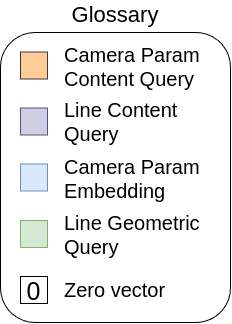}\\
	\end{tabular}}
	\resizebox{\linewidth}{!}{
		\begin{tabular}{cccc}
			\phantom{AAAAAA}a) CTRL-C \cite{ctrlc} & \phantom{AAAAA}b) CTRL-C w/ line content info & \phantom{AAAAa}c) SOFI (ours) & \phantom{aaaaaaaaAAAAAAAA}
	\end{tabular}}
	\caption{Decoder module. Query definition in different transformer-based model for camera calibration.}
	\label{fig: line_queries}
\end{figure}

We revisit the query formulation and how they are processed in the decoder in Fig. \ref{fig: line_queries}. CTRL-C \cite{ctrlc} and MSCC \cite{mscc}	initialize $\mathbf{q}_{\text{camera}}$ as in DETR \cite{detr}. However, $\mathbf{q}_{\text{line}}$ is initialized oppositely, as shown in Fig. \ref{fig: line_queries}a. This bad initialization for the line queries does not allow the line geometric information to propagate through the decoder layers. Additionally, their line query formulation does not allow the deformable attention mechanism \cite{deformable_detr} that promotes the cross-scale interaction.

First, we allow the propagation of the line geometric information as shown in Fig. \ref{fig: line_queries}b. We also added the line content information for a better understanding of the line. This simple and effective modification significantly boosts the performance of CTRL-C, as shown in Tab. \ref{tab: ctrl ablation}. Moreover, this new query initialization allows us to adapt deformable-DETR \cite{deformable_detr} for camera calibration tasks. Notice that with the zero vector initialization from Fig. \ref{fig: line_queries}a the reference all the line queries would share the same reference points, while ours allows different reference points for each line query.

\section{Loss function}
\subsection{Zenith Vanishing Point}
For the zenith vanishing point (zvp), given a ground-truth zvp $\mathbf{z}$ and a predicted zvp $\mathbf{\hat{z}}$, we define the loss between the two instances as:
\begin{equation}
	\mathcal{L}_{\text{zvp}} = 1 - \Bigg|\frac{\mathbf{z}^T\mathbf{\hat{z}}}{\|\mathbf{z} \|\|\mathbf{\hat{z}}\|}\Bigg|
\end{equation} 
where $\|\cdot\|$  represents the euclidean norm function. 
\subsection{Horizon Line}
Given a ground-truth horizon line ${\mathbf{hl}}$, 
we compute the left $\mathbf{b_l}$ and right $\mathbf{b_r}$ boundaries by intersecting $\mathbf{hl}$ to the image. We repeat the same procedure with the predicted horizon line $\mathbf{\hat{hl}}$ to compute $\mathbf{\hat{b}_l}$ and $\mathbf{\hat{b}_r}$. Then, we compute the loss for the horizon line as
\begin{equation}
	\mathcal{L}_{\text{hl}} = \mathrm{max}(\|\mathbf{b}_l -\mathbf{\hat{b}_l}\|_1,\|\mathbf{b}_r -\mathbf{\hat{b}_r}\|_1 ).
\end{equation}

\subsection{Field of View}
We define the loss function as:
\begin{equation}
	\mathcal{L}_{\text{fov}}= |f-\hat{f}|
\end{equation}
where $f$ and $\hat{f}$ correspond to the ground-truth and predicted field of view.
\subsection{Line Classification}
Following the procedure in \cite{ctrlc}, we produce pseudo ground truth for line classification. Given a line segment $\mathbf{l}$ and a vanishing point $\mathbf{v}$, the line intersects the vanishing point if $d(\mathbf{l}, \mathbf{v}) < \delta$, where $\delta = \mathrm{sin}(2^\circ)$, and $d(\cdot)$ is:
\begin{equation}
	d(\mathbf{l}, \mathbf{v}) =  \Bigg|\frac{\mathbf{l}^T\mathbf{v}}{\|\mathbf{l} \|\|\mathbf{v}\|}\Bigg|.
\end{equation}  

The Google Street dataset \cite{gsv} provides a zenith vanishing point $\mathbf{v}_{\text{z}}$ and two horizontal vanishing points $\mathbf{v}_{\text{hl1}}$ and $\mathbf{v}_{\text{hl2}}$. We group the lines depending on whether they pass through the zenith vanishing point or the horizon line with the following conditions:
\begin{align*}
	c(\mathbf{l}) = \begin{cases}
		2 & \text{if }  d(\mathbf{l}, \mathbf{v}_{\text{z}}) < \delta, 		\\
		1 & \text{if }  d(\mathbf{l}, \mathbf{v}_{\text{hl1}}) < \delta 	\text{ or }	d(\mathbf{l}, \mathbf{v}_{\text{hl2}}) < \delta,\\
		0 & \text{otherwise}.	
	\end{cases}
\end{align*}.

We also group the lines based on whether they pass through any vanishing point. We mathematically describe the process as follows:
\begin{align*}
	q(\mathbf{l}) = \begin{cases}
		1 & \text{if }  c(\mathbf{l}) = \{1, 2\}, 		\\
		0 & \text{otherwise}.	
	\end{cases}.
\end{align*}.

\section{More Ablation Studies}

\subsection{Effects of the New Loss Function}
CTRL-C \cite{ctrlc} and MSCC \cite{mscc} define the loss function as 
\begin{equation}
	\mathcal{L} = \mathcal{L}_{zvp} + \mathcal{L}_{hl} + \mathcal{L}_{fov} + \mathcal{L}_{verL} + \mathcal{L}_{horL}
\end{equation}
where all losses are weighted equally. However, they do not share the same importance. We follow DETR models \cite{deformable_detr, detr, dab} and give more importance to the main task ($\mathcal{L}_{zvp}$, $\mathcal{L}_{hl}$, and $\mathcal{L}_{fov}$). Our new loss function is
\begin{equation}
	\mathcal{L} = 5\mathcal{L}_{zvp} + 5\mathcal{L}_{hl} + 5\mathcal{L}_{fov} + \mathcal{L}_{verL} + \mathcal{L}_{horL}.
	\label{eq: new_loss}
\end{equation}

To validate our idea, we run CTRL-C and MSCC using the loss from equation \ref{eq: new_loss} and report the results in Table \ref{tab: new loss}. Training using eq. \ref{eq: new_loss} allows CTRL-C to have similar or better results than MSCC. In terms of the AUC, CTRL-C$^\dagger$ has slightly lower results. We believe this happens because detecting the horizon line has a stronger connection to the line classification task than the other task do. 

\begin{table*}[t]\centering
	\resizebox{\linewidth}{!}{%
		\begin{tabular}{lrcrcrcrcrc}
			\toprule
			Model &  & {Up ($^\circ$) $\downarrow$} & {\phantom{a}} & {Pitch ($^\circ$) $\downarrow$} & {\phantom{a}} & {Roll ($^\circ$) $\downarrow$} & {\phantom{a}} & {FoV ($^\circ$) $\downarrow$} & {\phantom{a}} & {AUC($\uparrow$)} \\
			\midrule
			CTRL-C \cite{ctrlc} && 1.80 && 1.58 && 0.66 && 3.59 && \textbf{87.29}\\
			MSCC \cite{mscc} && 1.72 && \textbf{1.50} && 0.62 && \textbf{3.21} && /\\
			CTRL-C$^{\dagger}$ && \textbf{1.71} && 1.52 && \textbf{0.57} && 3.38 && 87.16 \\
			\bottomrule
	\end{tabular}}
	\caption{Results on Google Street View Dataset \cite{gsv}. $\dagger$: means the model was trained using eq. \ref{eq: new_loss}.}
	\label{tab: new loss}
\end{table*}	

\subsection{CTRL-C with Line Content Information}
We conduct a study to validate the importance of combining line geometric and line content information for better scene and line understanding. We present the results in Table \ref{tab: ctrlc ablation}. Adding the line content information for the Google Street View dataset \cite{gsv} only slightly increases the model's accuracy. This may be due the fact that the dataset is relatively easy and the errors are already relatively low. On the other hand, we use the Holicity \cite{holicity} dataset for testing where there are significant improvements after adding each component. The most noticeable improvement is the AUC of the horizon line, which increases by almost 10 points compared to CTRL-C.

\begin{table*}[t]\centering
	\resizebox{\linewidth}{!}{%
		\begin{tabular}{lrcrcrcrcrc}
			\toprule
			Model &  & {Up ($^\circ$) $\downarrow$} & {\phantom{a}} & {Pitch ($^\circ$) $\downarrow$} & {\phantom{a}} & {Roll ($^\circ$) $\downarrow$} & {\phantom{a}} & {FoV ($^\circ$) $\downarrow$} & {\phantom{a}} & {AUC($\uparrow$)} \\
			\midrule
			&&\multicolumn{9}{c}{Google Street View \cite{gsv}}\\
			\cmidrule{3-11}
			CTRL-C && 1.80 && 1.58 && 0.66 && 3.59 && {87.29}\\
			\textcolor{blue}{+ eq. \ref{eq: new_loss}} && \textbf{1.71} && 1.52 && \textbf{0.57} && 3.38 && 87.16 \\
			\textcolor{blue}{+ line content} && \textbf{1.71} && \textbf{1.51} && 0.59 && \textbf{3.12} && \textbf{87.72} \\
			\cmidrule{3-11}
			&&\multicolumn{9}{c}{Holicity \cite{holicity}}\\
			\cmidrule{3-11}
			CTRL-C && 2.83 && 2.29 && 1.44 && 11.50 && 67.78\\
			\textcolor{blue}{+ eq. \ref{eq: new_loss}} && 2.66 && 2.26 && \textbf{1.09} && 12.41 && 72.31\\
			\textcolor{blue}{+ line content} && \textbf{2.55} && \textbf{2.13} && 1.14 && \textbf{11.46} && \textbf{77.57} \\				
			\bottomrule
	\end{tabular}}
	\caption{Ablation study of the different components added to CTRL-C \cite{ctrlc}.}
	\label{tab: ctrlc ablation}
\end{table*}	

\end{document}